\documentclass[conference]{IEEEtran}
\IEEEoverridecommandlockouts
\usepackage{cite}
\usepackage{amsmath,amssymb,amsfonts}
\usepackage{algorithmic}
\usepackage{graphicx}
\usepackage{textcomp}
\usepackage[acronym]{glossaries}
\usepackage{float}
\usepackage{stfloats}

\usepackage{xcolor}
\def\BibTeX{{\rm B\kern-.05em{\sc i\kern-.025em b}\kern-.08em
    T\kern-.1667em\lower.7ex\hbox{E}\kern-.125emX}}
\begin{document}

\title{HaT5: Hate Language Identification using Text-to-Text Transfer Transformer\\
}

\author{\IEEEauthorblockN{Sana Sabah Sabry\IEEEauthorrefmark{1},
Tosin Adewumi\IEEEauthorrefmark{2}, Nosheen Abid\IEEEauthorrefmark{3},
György Kovács\IEEEauthorrefmark{4}, Foteini Liwicki\IEEEauthorrefmark{5} \\and
Marcus Liwicki\IEEEauthorrefmark{6}}
\IEEEauthorblockA{Department of Computer Science, Electrical and Space Engineering\\
Luleå University of Technology\\
Luleå, Sweden \\
Email: \IEEEauthorrefmark{1}sana.sabah.sabry.al-azzawi@ltu.se,
\IEEEauthorrefmark{2}tosin.adewumi@ltu.se,
\IEEEauthorrefmark{3}nosheen.abid@ltu.set,
\IEEEauthorrefmark{4}gyorgy.kovacs@ltu.se,\\
\IEEEauthorrefmark{5}foteini.liwicki@ltu.se,
\IEEEauthorrefmark{6}marcus.liwicki@ltu.se
}}

\maketitle
\newacronym{nlp}{NLP}{Natural Language Processing}
\newacronym{ner}{NER}{Named Entity Recognition}
\newacronym{sa}{SA}{Sentiment Analysis}
\newacronym{bow}{BoW}{bag-of-words}
\newacronym{cbow}{CBoW}{continuous Bag-of-Words}
\newacronym{sltc}{SLTC}{Swedish Language Technology Conference}
\newacronym{ann}{ANN}{artificial neural network}
\newacronym{nn}{NN}{neural network}
\newacronym{lstm}{LSTM}{Long Short Term Memory Network}
\newacronym{sota}{SoTA}{state-of-the-art}
\newacronym{nlg}{NLG}{Natural Language Generation}
\newacronym{mwe}{MWE}{Multi-Word Expression}
\newacronym{cnn}{CNN}{Convolutional Neural Network}
\newacronym{sw}{SW}{Simple Wiki}
\newacronym{mt}{MT}{Machine Translation}
\newacronym{gdc}{GDC}{Gothenburg Dialog Corpus}
\newacronym{t5}{T5}{Text-to-Text-Transfer Transformer}
\newacronym{roberta}{RoBERTa}{Robustly optimized BERT approach}
\newacronym{bert}{BERT}{Bidirectional Encoder Representations from Transformers}
\newacronym{mcc}{MCC}{Matthews Correlation Coefficient}
\newacronym{ai}{AI}{Artificial Intelligence}
\newacronym{xai}{XAI}{explainable artificial intelligence}
\newacronym{lime}{LIME}{Local Interpretable Model-agnostic Explanations}
\newacronym{bilstm}{Bi-LSTM}{Bi-Directional Long Short Term Memory Network}
\newacronym{rnn}{RNN}{Recurrent Neural Network}
\newacronym{hasoc}{HASOC}{hate speech and offensive content}
\newacronym{olid}{OLID}{offensive language identification dataset}
\newacronym{ig}{IG}{Integrated Gradient}
\newacronym{relu}{ReLu}{Rectified Linear Unit}
 
\begin{abstract}
We investigate the performance of a \acrfull{sota} architecture \acrshort{t5} (available on the SuperGLUE) and compare with it 3 other previous \acrshort{sota} architectures across 5 different tasks from 2 relatively diverse datasets.
The datasets are diverse in terms of the number and types of tasks they have.
To improve performance, we augment the training data by using an autoregressive model.
We achieve near-\acrshort{sota} results on a couple of the tasks - macro F1 scores of 81.66\% for task A of the \acrshort{olid} 2019 dataset and 82.54\% for task A of the \acrfull{hasoc} 2021 dataset, where \acrshort{sota} are 82.9\% and 83.05\%, respectively.
We perform error analysis and explain why one of the models (\acrshort{bilstm}) makes the predictions it does by using a publicly available algorithm: \acrfull{ig}.
This is because \acrfull{xai} is essential for earning the trust of users.
The main contributions of this work are the implementation method of \acrshort{t5}, which is discussed; the data augmentation using a new conversational \acrshort{ai} model checkpoint, which brought performance improvements; and the revelation on the shortcomings of \acrshort{hasoc} 2021 dataset.
It reveals the difficulties of poor data annotation by using a small set of examples where the \acrshort{t5} model made the correct predictions, even when the ground truth of the test set were incorrect (in our opinion).
We also provide our model checkpoints on the HuggingFace hub\footnote{ https://huggingface.co/sana-ngu/HaT5\_augmentation \\ https://huggingface.co/sana-ngu/HaT5 \\ https://huggingface.co/sana-ngu/Hat5-Roberta} to foster transparency.
\end{abstract}


\section{Introduction}
Hate speech is usually viewed as any communication that disparages any person or group of people \cite{nockleby2000hate,brown2017hate}.
It is an unethical behaviour and has legal repercussions in many countries \cite{brown2017hate,quintel2020self}.
The increasing importance of detecting hate speech automatically has come to the fore with the increasing influence of social media, in addition to mainstream media.
For example, xenophobia and other social vices have been encouraged as a result of online/social media comments \cite{mutanga2020hate}.
Examples of hate speech, which may incite others in the \acrfull{olid} \cite{zampierietal2019} training data are given below.

\begin{itemize}
    \item @USER If Jamie Oliver fucks with my £3 meal deals at Tesco I’ll kill the cunt
    \item @USER She is wortless why is she not checked out by secret service she is irresponsible in trying to generate hste to have somone kill thre president
    \item @USER  Regarding the story I think I'm going to kill the president", why is it reporters "never" follow up on how a convicted felon could obtain, in this case, 6 guns? What's the police doing about, or would that bring light to the gun control issue? Bias?"
\end{itemize}

For societies to be peaceful and promote equality, it is crucial to (automatically) detect hate speech and address it.
This is also important to prevent perpetuating such undesirable behaviour or characteristics in new technologies, such as conversational systems and other AI technologies \cite{zhang2020dialogpt,adewumi2021sm}.
Automatic hate speech detection is imperative because of the herculean task of doing it manually and the delay that manual detection involves\footnote{bbc.com/news/world-europe-35105003}.
Besides, a manual approach is more likely to be subjective while a trained, automated approach is more objective \cite{javed2021understanding,adewumi2019conversational}. 

Our work spans 5 English tasks from 2 publicly available datasets.
The datasets are \acrshort{olid} \cite{zampierietal2019} and \acrshort{hasoc} 2021 \cite{mandl2021overview}.
The tasks are Tasks A \& B of the \acrshort{hasoc} 2021 and tasks A, B \& C of the \acrshort{olid} dataset.
Details of the datasets are provided in the Data section (\ref{datasets}).
These datasets were selected because of their importance and the subtasks covered with regards to hate speech.
This work compares the performance of different \acrshort{sota} architectures over these datasets.
The architectures include the \acrfull{bilstm}, the \acrfull{cnn}, \acrfull{roberta} and \acrfull{t5}.
We use pretrained models for both the \acrshort{roberta} and \acrshort{t5} from the HuggingFace library \cite{wolf-etal-2020-transformers}.

In addition, we perform data augmentation on the training set of \acrshort{hasoc} and evaluate the performance.
We investigate two types of data augmentation in this work and achieve near-\acrshort{sota} result on task A of the \acrshort{hasoc} dataset by using one of these approaches.
To promote transparency, we provide our model checkpoints for public access on the HuggingFace platform\footnote{ https://huggingface.co/sana-ngu/HaT5\_augmentation \\ https://huggingface.co/sana-ngu/HaT5 \\ https://huggingface.co/sana-ngu/Hat5-Roberta} and part of our codes\footnote{the \acrshort{t5} code will be released after a planned competition}.
Other key contributions of this paper are summarized as follows: 1) insight into the autoregressive data augmentation technique, 2) argument for more credible data annotation, and 3) investigation of the \acrshort{bilstm} with regards to explainability using \acrshort{ig} and error analysis for the \acrshort{t5} model.

The rest of the paper is divided as follows: section~\ref{related_work} provides an overview of prior work in automatic hate speech detection within the field of deep learning.
The datasets used in this work are described in section~\ref{datasets}.
Later, section~\ref{methodology} explains the methods used in this study.
Lastly, the results obtained, along with their critical analysis and discussion, are reported in section~\ref{results}.

\section{Related Work}
\label{related_work}

The prevalence of hate speech has probably increased with advancements in technology, making it easier and faster to spread on social media platforms.
This has motivated researchers to put significant efforts into creating datasets and designing intelligent algorithms for hate speech detection.
Linguists, in addition, have analyzed contents and defined different categories of hate speech data \cite{jaki2019online}.
Reference \cite{davidson2017automated} introduced the hate speech-offensive (HSO) dataset.
A logistic regression model with L2 regularization gave the best performance (F1 score of 0.9) during experimentation with the dataset.
In their approach, they trained the model on the entire dataset but used a one-versus-rest framework for prediction.

Reference \cite{mutanga2020hate} experimented with only the HSO dataset using different Transformer-based architectures.
Their preprocessing approach involved removing low frequency words from the dataset, though this may result in newly introduced terms of hate speech escaping detection.
Their best model (DistilBERT) achieved 0.75 F1 score.

The Transformer by \cite{vaswani2017attention} has gained fast adoption in the \acrshort{nlp} community since its introduction.
It is based solely on the attention mechanism and is, therefore, able to handle long-term dependencies \cite{bahdanau2015neural,vaswani2017attention}.
This advantage of models based on the attention mechanism makes dependency modeling possible regardless of distance in the sequence of input or output.
Therefore, the Transformer-based models have gained much attention in hate speech detection \cite{mutanga2020hate,matthew2020hatexplain,kovacs2021challenges}.
Among the many Transformer-based models are: \acrshort{bert} \cite{devlin-etal-2019-bert}, \acrshort{roberta} \cite{liu2019roberta} and \acrshort{t5} \cite{JMLR:v21:20-074}.

New datasets have been introduced since HSO while some have also been extended.
This shows the growing awareness of the importance of hate speech detection.
Besides extending the \acrshort{olid} dataset by \cite{zampierietal2019}, \cite{caselli2020feel} performed cross-domain experiments on HatEval \cite{basile2019semeval} after training on two different datasets.
Their reason for extending the OLID dataset was to annotate the distinction between messages deemed as explicit (containing slur or profanity) and implicit (having no slur).

\section{Data}
\label{datasets}
The following datasets are considered in this work:

\begin{enumerate}
    \item \acrshort{hasoc} 2021
    
    This is the third edition of \acrshort{hasoc} \cite{mandl2021overview}.
    Reference \cite{mandl2021overview} provided another set of tweets dataset with the same tasks as \acrshort{hasoc} 2020.
    The dataset includes the following subtasks: 1) Task A, which is to identify hate speech and offensive text and 2) Task B, which is a further classification for the previous task to categorize the hateful and offensive content into either hate content (HATE), offensive (OFFN) or profane (PRFN).
    The English dataset consists of 3,843 training samples and 1,281 samples in the test set.
    They gathered the data during the Covid-19 wave, therefore the text has Covid-related topics.
    We split 10\% of the training set as the dev (validation) set in this work for evaluation after each epoch
    
    \item OLID
    
    The dataset of OffensEval 2019 (Task 6 of SemEval 2019) is based on \acrshort{olid}.
    It contains 14,200 annotated English tweets and encompasses the following three subtasks: a) offensive language detection, b) categorization of offensive language as to whether it's targeted at someone (or a group) or not, and c) offensive language target identification such that distinction is made among individual, group and other entities, like an organisation or event \cite{zampierietal2019}.
    OLID is derived from Twitter. Crowdworkers performed its data annotation and the original data-split was into training and test sets only. Similarly with \acrshort{hasoc} 2021, we split 10\% of the training set as the dev (validation) set for evaluation after each epoch in this study.

\end{enumerate}

\section{Methodology}
\label{methodology}
The following subsections describe the different parts of the methodology.
All experiments were conducted on a shared DGX-1 machine with 8 × 32GB Nvidia V100 GPUs.
The server runs on Ubuntu 18 and has 80 CPU cores.
Each experiment was conducted 3 times and the average results (including standard deviation) obtained.
Each experiment was run for 6 epochs but the model checkpoint with the lowest validation loss is saved and used for evaluation of the test set.
Dev set results are also based on the model with the lowest validation loss.
We use a linear schedule with warm up for the learning rate adjustment for both \acrshort{roberta} and \acrshort{t5}.
We experimented with only very limited hyperparameters for all the models due to time and resource constraints.

\subsection{Models}
\subsubsection{\acrshort{bilstm}}
\acrshort{lstm} is one form of \acrfull{rnn}\cite{hochreiter1997long}. \acrshort{rnn} is used with sequential data and can capture long-term dependencies in text. \acrfull{bilstm} is another improved variant of \acrshort{rnn} that comprises of two \acrshort{lstm}s  where the input text flows forward and backwards, thereby providing more contextual information  and, as a result, improves the network performance \cite{graves2005framewise}.

First, we used an embedding layer to convert input text to its corresponding word embeddings.
Word and subword embeddings have been shown to improve performance of downstream tasks \cite{mikolov2013distributed,pennington2014glove,adewumi2020word2vec}.
We used Glove pre-trained word embedding \cite{pennington2014glove} of vector size 100 to capture the semantics of words with the surrounding context.
Then we applied a dropout layer to prevent overfitting.
We used 2 bi-directional layers; the dimension of the hidden state is 20. 
Finally, a fully connected layer follows the last \acrshort{bilstm} layer to classify the text.
This model has 1,317,721 parameters.

\subsubsection{\acrshort{cnn}}
\acrshort{cnn} was initially used for image processing. Reference \cite{kim-2014-convolutional} shows the effectiveness of the \acrshort{cnn} in capturing the local patterns in text on different \acrshort{nlp} tasks.  
Both the \acrshort{bilstm} and \acrshort{cnn} were used as feature-based models, where for each tweet, we computed embeddings using pre-trained Glove, then we used the sequence of embeddings as an input to the model.  
For the \acrshort{cnn}, the model is composed of 3  convolution layers with 100 filters each. The filter size for the first layer is 2*100 (100 is the size for the word embeddings), the filter size for the second layer is 3*100 and for the third layer it is 4*100.
We used a \acrfull{relu} activation function and max-pooling after each convolution layer.
Finally, we perform dropout on the outputs and fully connected layer for final classification.
The total trainable  parameters  for the CNN are 1,386,201

\subsubsection{\acrshort{roberta}}
\acrshort{roberta} is the product of a replication study of \acrshort{bert}.
It makes changes to \acrshort{bert} in the following ways: (1) training for longer over more data; (2) removing the next sentence prediction objective; (3) using longer sequences for training; and (4) changing the masking pattern dynamically when applied to the training data \cite{liu2019roberta}.
The base version of the model has 12 layers and 110M parameters.
For our study, we use \acrshort{roberta}-Base, a batch size of 32, initial learning rate of 1e-5 and maximum sequence length of 256.
We restricted the number of tasks to only binary tasks for this model.

\subsubsection{\acrshort{t5}}
The \acrshort{t5} \cite{JMLR:v21:20-074} follows the originally proposed transformer architecture by \cite{vaswani2017attention}.
It maps input sequence of tokens to embeddings before passing it to the encoder, comprising an alternating series of multi-head attention layer and feed-forward linear layer.
A different layer normalization is applied, where there's no additive bias applied and the activations are only rescaled.
The decoder includes a standard attention mechanism in addition to each self-attention layer.
Causal or autoregressive self-attention is used in the decoder for it to attend to past outputs.
Relative position embeddings are used instead of the original sinusoidal position signal, since self-attention is order-independent.
The \acrshort{t5}-Base model has about twice the number of parameters as that of \acrshort{bert}-Base.
Its training method uses teacher forcing (i.e. standard maximum likelihood) and a cross-entropy loss.
It was pretrained on 34B tokens, cutting the computational budget considerably when compared to \acrshort{bert} or \acrshort{roberta}, which were trained on 137B and 2.2T tokens, respectively.

Its base model has 220M parameters and 12 layers each in the encoder and decoder blocks \cite{JMLR:v21:20-074}.
Using this instead of the \acrshort{t5}-Small model required more memory and would not fit on a single V100 GPU for the batch size of 64, hence we lowered the batch size to 16.
The task prefix we use is `classification ' for all the tasks, as the model takes a hyperparameter called a task prefix. 

\subsection{Preprocessing}

Preprocessing was carried out on all the data to remove duplicates and unwanted strings or characters.
For example, for OLID (task C), there are nans (empty entries) in some columns of the labels.
The nans cause problems for the models, increasing the total number of categories and dropping model performance.
We, therefore, dropped all such rows during the preprocessing step.
Furthermore, tweets are noisy and unstructured, which can affect the performance of the models.
To prepare the text for each model, the following preprocessing steps were applied to the datasets: 

\begin{itemize} 

\item URLs are excluded. 

\item Uppercase  characters are changed to lowercase. 

\item Emails are removed.    

\item Spaces more than 2 characters long are replaced as one space.

\item Special characters such as hashtags(\#), emojis and mention symbols (\makeatletter @)  are removed. 

\item Numbers are removed. 

\item IP addresses are removed

\end{itemize} 

\subsection{Data Augmentation}
We experimented with two data augmentation techniques: 1) a simple token-level deletion of the start and end tokens for each sample and 2) autoregressive text generation using the model checkpoint by \cite{adewumi2021sm}.
It was fine-tuned on the MultiWOZ dataset by \cite{eric-EtAl:2020:LREC}.
The autoregressive model is originally a pretrained DialoGPT (medium) model \cite{zhang2020dialogpt}.
The second technique effectively doubled the training set size.
In the first technique, samples ending or starting with offensive words are kept as they are in the new augmented training data and are therefore dropped when merged with the original, to avoid duplicates.
Examples from HASOC 2021 of the original sample and the augmented sample, using the second technique, are given below.

\begin{itemize}
    \item Bloody hell is that all they can do ? https://t.co/JQOiOw8nPs
    \item bloody hell is that all they can do It is. Have a great day.\\
    \item He voted against migration by voting brexit the wanker https://t.co/5t419W0iq9
    \item he voted against migration by voting brexit the wanker I'm confused. Are you saying you don't have a single moderate priced hotel in the centre of town with free parking?\\
    \item SHOOT NOW ASSHOLE
    \item shoot now asshole Booking was successful. Reference number is : N0LQRA43.\\
\end{itemize}

For the first technique, we use the list of offensive words available from the online resource at Carnegie Mellon University\footnote{cs.cmu.edu/~biglou/resources/} to ensure offensive or hate speech tokens are not dropped.
The document originally contains almost 1,400 English terms that could be found offensive.
However, we removed some words (160), which are nationalities/geographical locations or adjectives of emotions.
We consider that these may not qualify as offensive words.
Examples of such words are: african, american, arab, canadian, european, angry and many non-harmful words.
There are 1,223 words left in the document we used for our experiment.
The first technique was found to be less effective in improving model performance so we did not report its results in the next section.

\subsection{Metrics}
The F1 score is the harmonic mean of the precision and recall.
The relative contribution to the F1 from precision and recall are equal.
Micro F1 calculates globally by counting the total true positives, false negatives and false positives.
Macro F1 does not take label imbalance into account unlike weighted F1, which accounts for label imbalance by finding the labels' mean weighted by support (each label's true instances) \cite{scikit-learn}\footnote{scikit-learn.org/stable/modules/generated/sklearn.metrics.f1\_score.html}.
We report both weighted and macro F1 scores because of past studies' reported metrics.

\section{Results and Discussion}
\label{results}
Table \ref{tab:compare} shows the results obtained across the models and datasets, and those from \cite{zampieri2019semeval} and \cite{mandl2021overview}.
We report both weighted and macro F1 scores for the dev and test sets, where applicable.
Also, the standard deviation (sd) is reported.
Considering the test set results, it is perhaps not surprising that the Transformer-based models outperform both the \acrshort{bilstm} and \acrshort{cnn} models, in almost all instances.
The \acrshort{bilstm} obtains better results than the \acrshort{cnn} in the tasks A of both \acrshort{hasoc} and \acrshort{olid} but the \acrshort{cnn} outperforms it with a smaller margin in the tasks B \% C of \acrshort{olid}.
It also outperforms the \acrshort{bilstm} in task B of \acrshort{hasoc}.
\acrshort{t5} outperforms \acrshort{roberta} on all tasks.
It has near-\acrshort{sota} performance in task A of \acrshort{olid}, given the result in \cite{zampieri2019semeval}.
The \acrshort{t5}+Augmented Data version shows improved scores in both tasks of \acrshort{hasoc} when compared with the plain \acrshort{t5}.
This makes the result near-\acrshort{sota} when compared with \cite{mandl2021overview}.
Adequate exploration of hyperparameter tuning may have produced better scores with the models but we did not have sufficient time to do this.

\begin{table}[h!]
\caption{\label{tab:compare} Model comparison of mean scores for OLID \& HASOC tasks. \footnotesize{(sd: standard deviation; bold values are best scores for a given task; '-' implies no informaton available)}}
\centering
\resizebox{\columnwidth}{!}{%
\begin{tabular}{lcccc}
\hline
\textbf{Task} &
\multicolumn{2}{c}{\textbf{Weighted F1 (\%)}} &
\multicolumn{2}{c}{\textbf{Macro F1(\%)}}
\\
 & Dev (sd) & Test (sd) & Dev (sd) & Test (sd)
\\
\hline
\textbf{LSTM} & \multicolumn{4}{c}{}
\\
\hline
OLID A &  79.59 (0.89) & 83.89 (0.57) &  78.48 (1.52) & 79.49 (0)
\\
OLID B &  82.50 (1.70) & 83.46 (0) &  46.76 (0) & 47.32 (0)
\\
OLID C &  49.75 (3.95) & 43.82 (9.63) &  35.65 (2.81) & 36.82 (0)
\\
Hasoc 2021 A &  78.05 (0.85) &  78.43 (0.84) &  77.99 (1.79) &  77.19 (0)
\\
Hasoc 2021 B &  50.65 (1.34) &  52.19 (1.95) & 43.19 (2.09) & 42.25 (0)
\\
\hline
\textbf{CNN} & \multicolumn{4}{c}{}
\\
\hline
OLID A &  79.10 (0.26) & 82.47 (0.56)  &  77.61 (0.39) & 78.46 (0)
\\
OLID B &  82.43 (0.49) & 83.46 (0) & 46.76 (0) & 47.88 (0)
\\
OLID C &  47.54 (1.36) & 38.09 (3.91) &  35.65 (0) & 36.85 (0)
\\
Hasoc 2021 A &  77.22 (0.52) &  77.63 (0.70) &   74.28 (0.58) &   75.67 (0)
\\
Hasoc 2021 B &  55.60 (0.61) &  59.84 (0.41) & 50.41 (0.41) & 54.99 (0)
\\
\hline
\textbf{RoBERTa} & \multicolumn{4}{c}{}
\\
\hline
OLID A & 82.70 (0.55) & 84.62 (0) &  80.51 (0.76) & 80.34 (0)
\\
OLID B & 82.70 (0.13) & 83.46 (0) & 46.76 (0.04) & 47.02 (0)
\\
Hasoc 2021 A & 79.9 (0.57) & 76.2 (0) & 77.77 (0.75) & 74 (0)
\\
\hline
\textbf{T5-Base} & \multicolumn{4}{c}{}
\\
\hline
OLID A & 92.90 (1.37) & 85.57 (0) & 92.93 (1.42) & 81.66 (0)
\\
OLID B & 99.75 (0.43) & 86.81 (0) & 99.77 (0.44) & 53.78 (0)
\\
OLID C & 58.35 (1.22) & 54.99 (0) & 33.09 (0.76) & 43.12 (0)
\\
Hasoc 2021 A & 94.60 (1.98) & 82.3 (0) & 94.73 (5.26) & 80.81 (0)
\\
Hasoc 2021 B & 65.40 (0.82) & 62.74 (0) & 62.43 (6.32) & 59.21 (0)
\\
\hline
\textbf{T5-Base+Augmented Data} & \multicolumn{4}{c}{}
\\
\hline
Hasoc 2021 A & 95.5 (3.27) & 83 (0) & 92.97 (2.20) & 82.54 (0)
\\
Hasoc 2021 B & 64.74 (3.84) & 66.85 (0) & 65.56 (1.48) & 62.71 (0)
\\
\hline
\textbf{\cite{zampieri2019semeval}} best scores& \multicolumn{4}{c}{}
\\
\hline
OLID A &  & &  & \textbf{82.90} (-)
\\
OLID B &  &  &  & \textbf{75.50} (-)
\\
OLID C &  &  & & \textbf{66} (-)
\\
\hline
\textbf{\cite{mandl2021overview}} best scores & \multicolumn{4}{c}{\textbf{}}
\\
\hline
Hasoc 2021 A &  &  &  &  \textbf{83.05} (-)
\\
Hasoc 2021 B &  &  &  & \textbf{66.57} (-)
\\
\hline
\end{tabular}
}
\end{table}

We observe the performance of the \acrshort{lstm} trails behind that of the \acrshort{t5}.
Indeed, the \acrshort{t5} may have performed even better but for a certain shortcoming.
Text classification in \acrshort{t5} outputs a prediction of a single word of the target label.
However, in our experiments, we found out that the model is more stable with predictions when fine-tuned with target labels of numbers, explicitly type-cast as string.
Otherwise, some predictions during fine-tuning can be an empty string or expressions from the training set, especially in the early epochs of the training.
This problem is noted as a possibility by \cite{JMLR:v21:20-074} though they did not experience this behaviour in their trained models.
Using target labels of numbers, explicitly type-cast as string greatly reduces this occurrence.

\subsection{Error Analysis}
\begin{figure}[h]
\centering
\includegraphics[width=0.4\textwidth]{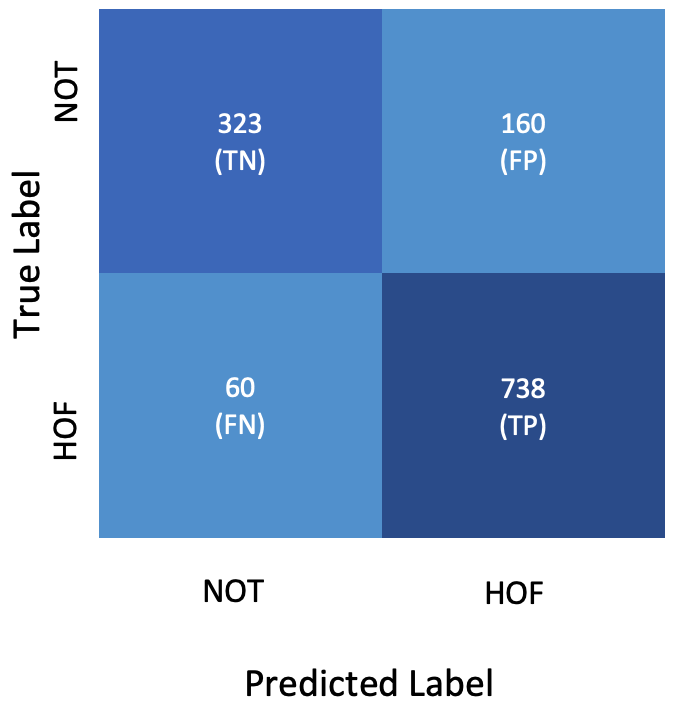}
\caption{Confusion matrix of \acrshort{t5} on Hasoc 2021 test set}
\label{fig:cmat}
\end{figure}

We investigate the errors made by the \acrshort{t5} model on the HASOC 2021 test set.
Figure \ref{fig:cmat} reveals the distribution of the predictions in a confusion matrix.
33.13\% (160) of the `NOT' class (not offensive) was misclassified while only 7.52\% of the `HOF' (hate or offensive) class was misclassified.
This may not be unconnected to the fact that the training set had more `HOF' samples, so the model is better at identifying such.
This is because the training set has 2,251 `HOF' samples and 1,207 `NOT' samples.
Improving the model's ability on identifying `NOT' samples (and the overall performance) may involve using a balanced training set or stratifying the categories during training.

A strong case for better annotation of data is evident when one considers some interesting cases in the HASOC 2021 test set.
The examples below were annotated (with the ground truth) of being not offensive (NOT), however, the model was intelligent enough to predict them as hate or offensive (HOF).
This issue has implications for the assessments done during the competitions organized.
Furthermore, it is certain the \acrshort{t5} and other models would have reported better scores if not for the issue raised here.


\begin{table}[h!]
\caption{Examples from HASOC 2021 test set with suspicious labels}
\centering
\resizebox{\columnwidth}{!}{%
\begin{tabular}{p{0.25\linewidth} p{0.55\linewidth}p{0.10\linewidth}p{0.10\linewidth}}
\hline
\_id & text & task\_1 & task\_2
\\
\hline
60c5d6bf5659 ea5e55df0242 & miya four creeps into every thought i have what the fuck & NOT & NONE 
\\
\hline
60c5d6bf5659 ea5e55defe58 & At least we’re being freed from the shambles of the evangelical, but damn y’all couldn’t just go w the flow? Y’all just had to find jebus through crystals and astrology? Smdh & NOT & NONE 
\\ 
\hline
60c5d6bf5659 ea5e55defb73 & These terrorists are more dangerous than \#chinesevirus https:\/\/t.co\/iazlTYm8St & NOT & NONE 
\\
\hline
60c5d6bf5659 ea5e55defe24 & @kumarmbayar @Actor\_Siddharth He is mentally Upset, Why he's not talking about \#BengalBurning and why stupid talks about \@Tejasvi\_Surya & NOT & NONE
\\
\hline
\end{tabular}
}
\label{table:roberta}
\end{table}

\begin{figure*}[bp]
\centering
\includegraphics[width=0.9\textwidth]{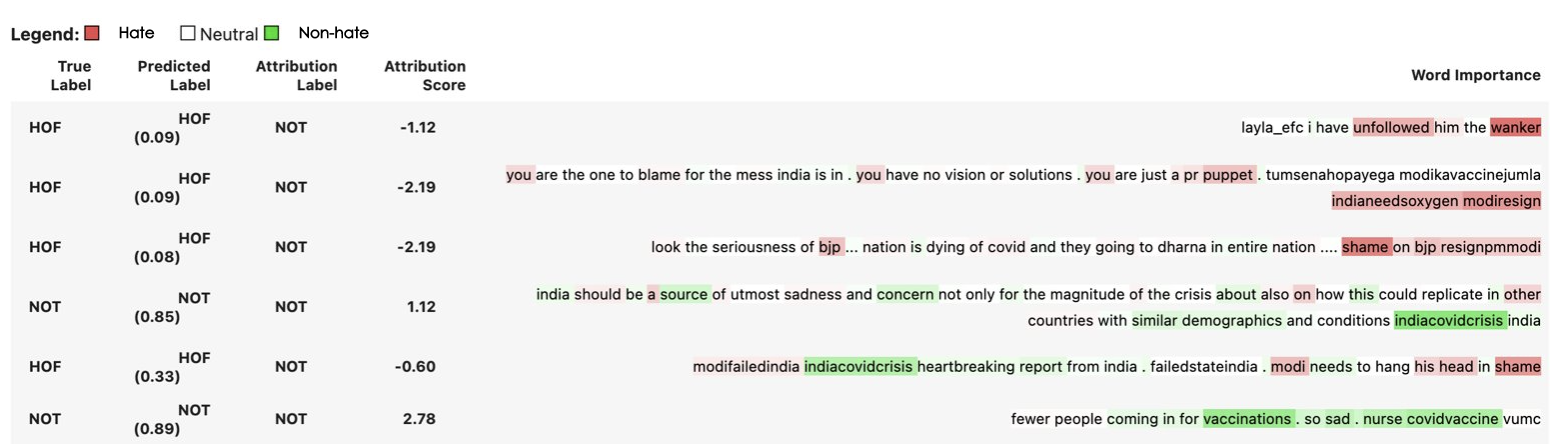}
\caption{Visualize attributions for \acrshort{bilstm} on \acrshort{hasoc} 2021 test set (correct-classification)}
\label{fig:xai1}
\end{figure*}

\begin{figure*}[t]
\centering
\includegraphics[width=0.9\textwidth]{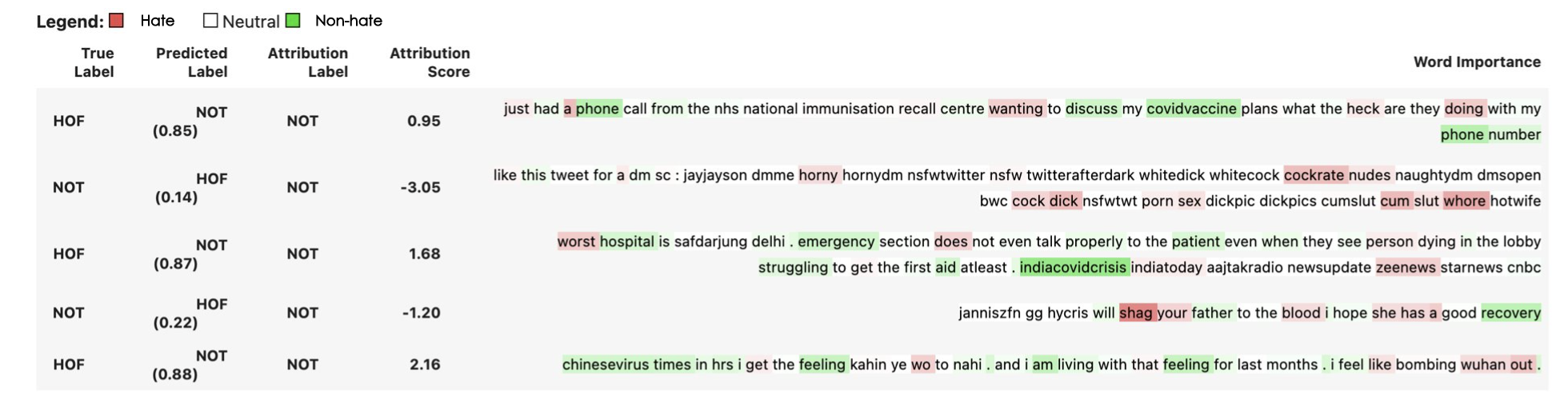}
\caption{Visualize attributions for \acrshort{bilstm} on Hasoc 2021 test set (miss-classification)}
\label{fig:xai2}
\end{figure*}

\subsection{Explanability (XAI)}
Explainable \acrshort{ai} (\acrshort{xai}) is essential for trust, and the purpose of this section is to understand why the \acrshort{bilstm}  model miss-classified specific tweets and what makes the model struggle in making the correct prediction.  
There are many ways to explain NLP models, and several researchers have proposed different methods to explain the output of the ML black box \cite{ribeiro2016should} \cite{lundberg2017unified}. 
In this study, we use \acrfull{ig}.
\acrshort{ig} is a simple but powerful axiomatic attribution originally proposed by \cite{sundararajan2017axiomatic}. 
It is an attribution method that attributes a model’s prediction to the input features (attributing the text classification to individual words in this case).  
We have computed the attribution compared to a baseline sequence of tokens by creating a path from the baseline to the input text.
At each step on that path, we aggregate the gradient and finally, we calculate the path integral for the aggregated gradients.
The attribution showed which word of the text input to the model affects the model prediction and how strongly.

In both Figures \ref{fig:xai1} and \ref{fig:xai2}, the important words are highlighted.
The words in green contribute to non-hate speech; those highlighted in red contribute to hate speech. 
Figure \ref{fig:xai2} shows examples of wrongly classified tweets.
The second tweet has offensive words but it is annotated wrongly as `NOT' (not offensive).
The \acrshort{bilstm}, however, predicts this correctly.
In the third example, the model makes the prediction `NOT' mainly because of the hashtag ‘indiacovidcrisis’, which is strongly associated with non-hate in both figures.
This is very likely because, in the training set, most of the tweets containing this hashtag are classified as non-hateful content.
In the fourth tweet, the most important words to the model for considering the tweet as hateful are ‘shag’ and ‘blood’.
However, the model may have misunderstood the context of the tweet.
For the last example, the model appears completely oblivious to the word `bombing' and paid attention to other words, which ended in its wrong prediction. 

\section{Conclusion}
Automatically detecting hate speech is a very important task and we show in this study that progress has been made, especially with the Transformer-based architectures.
We also show that the quality of annotated data is crucial for the success of automatically detecting hate speech (and other offensive communication).
We compared the performance of different \acrshort{sota} architectures over multiple tasks in 2 datasets.
The \acrshort{t5} pretrained model outperformed the \acrshort{lstm}, \acrshort{cnn} and \acrshort{roberta} architectures/models.
Data augmentation provided additional performance gains, establishing near-\acrshort{sota} result on tasks A of the \acrshort{hasoc} 2021 datasets while the plain \acrshort{t5} achieved near-\acrshort{sota} performance on task A of \acrshort{olid} 2019.

Future direction may include using
a voting ensemble method.
Although this is potentially powerful, it may suffer due to poor votes from weak models.
Another direction is improving the predictive power of models in cross-domain or zero-shot inference.
Finally, it is also important to study the performance of models on raw data, even though preprocessing is an important step in obtaining better performance \cite{obaid2019impact}.
This is because some offensive messages may contain only disparaging emojis or special characters or other expressions of hate speech and not text.
\bibliographystyle{IEEEtran}
\bibliography{ref}

\begin{thebibliography}{10}
\providecommand{\url}[1]{#1}
\csname url@samestyle\endcsname
\providecommand{\newblock}{\relax}
\providecommand{\bibinfo}[2]{#2}
\providecommand{\BIBentrySTDinterwordspacing}{\spaceskip=0pt\relax}
\providecommand{\BIBentryALTinterwordstretchfactor}{4}
\providecommand{\BIBentryALTinterwordspacing}{\spaceskip=\fontdimen2\font plus
\BIBentryALTinterwordstretchfactor\fontdimen3\font minus
  \fontdimen4\font\relax}
\providecommand{\BIBforeignlanguage}[2]{{%
\expandafter\ifx\csname l@#1\endcsname\relax
\typeout{** WARNING: IEEEtran.bst: No hyphenation pattern has been}%
\typeout{** loaded for the language `#1'. Using the pattern for}%
\typeout{** the default language instead.}%
\else
\language=\csname l@#1\endcsname
\fi
#2}}
\providecommand{\BIBdecl}{\relax}
\BIBdecl

\bibitem{nockleby2000hate}
J.~T. Nockleby, ``Hate speech,'' \emph{Encyclopedia of the American
  constitution}, vol.~3, no.~2, pp. 1277--1279, 2000.

\bibitem{brown2017hate}
A.~Brown, ``What is hate speech? part 1: The myth of hate,'' \emph{Law and
  Philosophy}, vol.~36, no.~4, pp. 419--468, 2017.

\bibitem{quintel2020self}
T.~Quintel and C.~Ullrich, ``Self-regulation of fundamental rights? the eu code
  of conduct on hate speech, related initiatives and beyond,'' in
  \emph{Fundamental Rights Protection Online}.\hskip 1em plus 0.5em minus
  0.4em\relax Edward Elgar Publishing, 2020.

\bibitem{mutanga2020hate}
R.~Mutanga, N.~Naicker, and O.~O. Olugbara, ``Hate speech detection in twitter
  using transformer methods,'' \emph{International Journal of Advanced Computer
  Science and Applications}, vol.~11, no.~01, 2020.

\bibitem{zampierietal2019}
M.~Zampieri, S.~Malmasi, P.~Nakov, S.~Rosenthal, N.~Farra, and R.~Kumar,
  ``{Predicting the Type and Target of Offensive Posts in Social Media},'' in
  \emph{Proceedings of NAACL}, 2019.

\bibitem{zhang2020dialogpt}
Y.~{Zhang}, S.~{Sun}, M.~{Galley}, Y.-C. {Chen}, C.~{Brockett}, X.~{Gao},
  J.~{Gao}, J.~{Liu}, and B.~{Dolan}, ``Dialogpt: Large-scale generative
  pre-training for conversational response generation,'' in \emph{Proceedings
  of the 58th Annual Meeting of the Association for Computational Linguistics:
  System Demonstrations}, 2020, pp. 270--278.

\bibitem{adewumi2021sm}
T.~Adewumi, N.~Abid, M.~Pahlavan, R.~Br{\"a}nnvall, S.~S. Sabry, F.~Liwicki,
  and M.~Liwicki, ``Sm $\{$$\backslash$aa$\}$ prat: Dialogpt for natural
  language generation of swedish dialogue by transfer learning,'' \emph{arXiv
  preprint arXiv:2110.06273}, 2021.

\bibitem{javed2021understanding}
S.~Javed, T.~P. Adewumi, F.~S. Liwicki, and M.~Liwicki, ``Understanding the
  role of objectivity in machine learning and research evaluation,''
  \emph{Philosophies}, vol.~6, no.~1, p.~22, 2021.

\bibitem{adewumi2019conversational}
T.~P. Adewumi, F.~Liwicki, and M.~Liwicki, ``Conversational systems in machine
  learning from the point of view of the philosophy of science—using alime
  chat and related studies,'' \emph{Philosophies}, vol.~4, no.~3, p.~41, 2019.

\bibitem{mandl2021overview}
T.~Mandl, S.~Modha, G.~K. Shahi, H.~Madhu, S.~Satapara, P.~Majumder,
  J.~Schaefer, T.~Ranasinghe, M.~Zampieri, D.~Nandini \emph{et~al.}, ``Overview
  of the hasoc subtrack at fire 2021: Hate speech and offensive content
  identification in english and indo-aryan languages,'' \emph{arXiv preprint
  arXiv:2112.09301}, 2021.

\bibitem{wolf-etal-2020-transformers}
\BIBentryALTinterwordspacing
T.~Wolf, L.~Debut, V.~Sanh, J.~Chaumond, C.~Delangue, A.~Moi, P.~Cistac,
  T.~Rault, R.~Louf, M.~Funtowicz, J.~Davison, S.~Shleifer, P.~von Platen,
  C.~Ma, Y.~Jernite, J.~Plu, C.~Xu, T.~Le~Scao, S.~Gugger, M.~Drame, Q.~Lhoest,
  and A.~Rush, ``Transformers: State-of-the-art natural language processing,''
  in \emph{Proceedings of the 2020 Conference on Empirical Methods in Natural
  Language Processing: System Demonstrations}.\hskip 1em plus 0.5em minus
  0.4em\relax Online: Association for Computational Linguistics, Oct. 2020, pp.
  38--45. [Online]. Available:
  \url{https://aclanthology.org/2020.emnlp-demos.6}
\BIBentrySTDinterwordspacing

\bibitem{jaki2019online}
S.~Jaki, T.~De~Smedt, M.~Gw{\'o}{\'z}d{\'z}, R.~Panchal, A.~Rossa, and
  G.~De~Pauw, ``Online hatred of women in the incels. me forum: Linguistic
  analysis and automatic detection,'' \emph{Journal of Language Aggression and
  Conflict}, vol.~7, no.~2, pp. 240--268, 2019.

\bibitem{davidson2017automated}
T.~Davidson, D.~Warmsley, M.~Macy, and I.~Weber, ``Automated hate speech
  detection and the problem of offensive language,'' in \emph{Proceedings of
  the International AAAI Conference on Web and Social Media}, vol.~11, no.~1,
  2017.

\bibitem{vaswani2017attention}
A.~Vaswani, N.~Shazeer, N.~Parmar, J.~Uszkoreit, L.~Jones, A.~N. Gomez,
  {\L}.~Kaiser, and I.~Polosukhin, ``Attention is all you need,'' in
  \emph{Advances in neural information processing systems}, 2017, pp.
  5998--6008.

\bibitem{bahdanau2015neural}
\BIBentryALTinterwordspacing
D.~Bahdanau, K.~Cho, and Y.~Bengio, ``Neural machine translation by jointly
  learning to align and translate,'' in \emph{International Conference on
  Learning Representations, {ICLR} 2015}, 2015. [Online]. Available:
  \url{https://arxiv.org/pdf/1409.0473.pdf}
\BIBentrySTDinterwordspacing

\bibitem{matthew2020hatexplain}
B.~Mathew, P.~Saha1, S.~M. Yimam, C.~Biemann, P.~Goyal1, and A.~Mukherjee1,
  ``Hatexplain: A benchmark dataset for explainable hate speech detection,'' in
  \emph{Proceedings of the 35th Association for the Advancement of Artificial
  Intelligence Conference on Artificial Intelligence}, 2021.

\bibitem{kovacs2021challenges}
G.~Kov{\'a}cs, P.~Alonso, and R.~Saini, ``Challenges of hate speech detection
  in social media,'' \emph{SN Computer Science}, vol.~2, no.~2, pp. 1--15,
  2021.

\bibitem{devlin-etal-2019-bert}
\BIBentryALTinterwordspacing
J.~Devlin, M.-W. Chang, K.~Lee, and K.~Toutanova, ``{BERT}: Pre-training of
  deep bidirectional transformers for language understanding,'' in
  \emph{Proceedings of the 2019 Conference of the North {A}merican Chapter of
  the Association for Computational Linguistics: Human Language Technologies,
  Volume 1 (Long and Short Papers)}.\hskip 1em plus 0.5em minus 0.4em\relax
  Minneapolis, Minnesota: Association for Computational Linguistics, Jun. 2019,
  pp. 4171--4186. [Online]. Available: \url{https://aclanthology.org/N19-1423}
\BIBentrySTDinterwordspacing

\bibitem{liu2019roberta}
Y.~Liu, M.~Ott, N.~Goyal, J.~Du, M.~Joshi, D.~Chen, O.~Levy, M.~Lewis,
  L.~Zettlemoyer, and V.~Stoyanov, ``Roberta: A robustly optimized bert
  pretraining approach,'' \emph{arXiv preprint arXiv:1907.11692}, 2019.

\bibitem{JMLR:v21:20-074}
\BIBentryALTinterwordspacing
C.~Raffel, N.~Shazeer, A.~Roberts, K.~Lee, S.~Narang, M.~Matena, Y.~Zhou,
  W.~Li, and P.~J. Liu, ``Exploring the limits of transfer learning with a
  unified text-to-text transformer,'' \emph{Journal of Machine Learning
  Research}, vol.~21, no. 140, pp. 1--67, 2020. [Online]. Available:
  \url{http://jmlr.org/papers/v21/20-074.html}
\BIBentrySTDinterwordspacing

\bibitem{caselli2020feel}
T.~Caselli, V.~Basile, J.~Mitrovi{\'c}, I.~Kartoziya, and M.~Granitzer, ``I
  feel offended, don’t be abusive! implicit/explicit messages in offensive
  and abusive language,'' in \emph{Proceedings of the 12th language resources
  and evaluation conference}, 2020, pp. 6193--6202.

\bibitem{basile2019semeval}
V.~Basile, C.~Bosco, E.~Fersini, N.~Debora, V.~Patti, F.~M.~R. Pardo, P.~Rosso,
  M.~Sanguinetti \emph{et~al.}, ``Semeval-2019 task 5: Multilingual detection
  of hate speech against immigrants and women in twitter,'' in \emph{13th
  International Workshop on Semantic Evaluation}.\hskip 1em plus 0.5em minus
  0.4em\relax Association for Computational Linguistics, 2019, pp. 54--63.

\bibitem{hochreiter1997long}
S.~Hochreiter and J.~Schmidhuber, ``Long short-term memory,'' \emph{Neural
  computation}, vol.~9, no.~8, pp. 1735--1780, 1997.

\bibitem{graves2005framewise}
A.~Graves and J.~Schmidhuber, ``Framewise phoneme classification with
  bidirectional lstm and other neural network architectures,'' \emph{Neural
  networks}, vol.~18, no. 5-6, pp. 602--610, 2005.

\bibitem{mikolov2013distributed}
T.~Mikolov, I.~Sutskever, K.~Chen, G.~S. Corrado, and J.~Dean, ``Distributed
  representations of words and phrases and their compositionality,'' in
  \emph{Advances in neural information processing systems}, 2013, pp.
  3111--3119.

\bibitem{pennington2014glove}
J.~Pennington, R.~Socher, and C.~D. Manning, ``Glove: Global vectors for word
  representation,'' in \emph{Proceedings of the 2014 conference on empirical
  methods in natural language processing (EMNLP)}, 2014, pp. 1532--1543.

\bibitem{adewumi2020word2vec}
T.~P. Adewumi, F.~Liwicki, and M.~Liwicki, ``Word2vec: Optimal hyper-parameters
  and their impact on nlp downstream tasks,'' \emph{arXiv preprint
  arXiv:2003.11645}, 2020.

\bibitem{kim-2014-convolutional}
\BIBentryALTinterwordspacing
Y.~Kim, ``Convolutional neural networks for sentence classification,'' in
  \emph{Proceedings of the 2014 Conference on Empirical Methods in Natural
  Language Processing ({EMNLP})}.\hskip 1em plus 0.5em minus 0.4em\relax Doha,
  Qatar: Association for Computational Linguistics, Oct. 2014, pp. 1746--1751.
  [Online]. Available: \url{https://aclanthology.org/D14-1181}
\BIBentrySTDinterwordspacing

\bibitem{eric-EtAl:2020:LREC}
\BIBentryALTinterwordspacing
M.~Eric, R.~Goel, S.~Paul, A.~Sethi, S.~Agarwal, S.~Gao, A.~Kumar, A.~Goyal,
  P.~Ku, and D.~Hakkani-Tur, ``Multiwoz 2.1: A consolidated multi-domain
  dialogue dataset with state corrections and state tracking baselines,'' in
  \emph{Proceedings of The 12th Language Resources and Evaluation
  Conference}.\hskip 1em plus 0.5em minus 0.4em\relax Marseille, France:
  European Language Resources Association, May 2020, pp. 422--428. [Online].
  Available: \url{https://www.aclweb.org/anthology/2020.lrec-1.53}
\BIBentrySTDinterwordspacing

\bibitem{scikit-learn}
F.~Pedregosa, G.~Varoquaux, A.~Gramfort, V.~Michel, B.~Thirion, O.~Grisel,
  M.~Blondel, P.~Prettenhofer, R.~Weiss, V.~Dubourg, J.~Vanderplas, A.~Passos,
  D.~Cournapeau, M.~Brucher, M.~Perrot, and E.~Duchesnay, ``Scikit-learn:
  Machine learning in {P}ython,'' \emph{Journal of Machine Learning Research},
  vol.~12, pp. 2825--2830, 2011.

\bibitem{zampieri2019semeval}
M.~Zampieri, S.~Malmasi, P.~Nakov, S.~Rosenthal, N.~Farra, and R.~Kumar,
  ``Semeval-2019 task 6: Identifying and categorizing offensive language in
  social media (offenseval),'' \emph{arXiv preprint arXiv:1903.08983}, 2019.

\bibitem{ribeiro2016should}
M.~T. Ribeiro, S.~Singh, and C.~Guestrin, ``" why should i trust you?"
  explaining the predictions of any classifier,'' in \emph{Proceedings of the
  22nd ACM SIGKDD international conference on knowledge discovery and data
  mining}, 2016, pp. 1135--1144.

\bibitem{lundberg2017unified}
S.~M. Lundberg and S.-I. Lee, ``A unified approach to interpreting model
  predictions,'' in \emph{Proceedings of the 31st international conference on
  neural information processing systems}, 2017, pp. 4768--4777.

\bibitem{sundararajan2017axiomatic}
M.~Sundararajan, A.~Taly, and Q.~Yan, ``Axiomatic attribution for deep
  networks,'' in \emph{International Conference on Machine Learning}.\hskip 1em
  plus 0.5em minus 0.4em\relax PMLR, 2017, pp. 3319--3328.

\bibitem{obaid2019impact}
H.~S. Obaid, S.~A. Dheyab, and S.~S. Sabry, ``The impact of data pre-processing
  techniques and dimensionality reduction on the accuracy of machine
  learning,'' in \emph{2019 9th Annual Information Technology,
  Electromechanical Engineering and Microelectronics Conference
  (IEMECON)}.\hskip 1em plus 0.5em minus 0.4em\relax IEEE, 2019, pp. 279--283.

\end{thebibliography}

\vspace{12pt}

\printglossary[type=\acronymtype]

\end{document}